\documentclass[11pt,a4paper]{article}
\usepackage[hyperref]{emnlp2020}
\usepackage{times}
\usepackage{latexsym}
\usepackage{mathtools,float,amssymb,graphicx}
\usepackage{makecell}

\usepackage{microtype}
\usepackage{enumitem}
\setlist{itemsep=0px,parsep=0px,topsep=0.5\baselineskip,leftmargin=10px}

\usepackage[utf8]{inputenc}
\usepackage{multirow}

\usepackage{booktabs}

\aclfinalcopy 
\def\aclpaperid{3116} 

\newcommand{\commentfn}[1]{}
\renewcommand{\commentfn}[1]{#1} 

\newcommand{\tydi}{\textsc{TyDi QA}}
\newcommand{\minspan}{\textsc{MinSpan}}
\newcommand{\selectp}{\textsc{SelectP}}
\newcommand{\avg}{\textbf{Avg}}
\newcommand{\joint}{\mbox{\textsc{Joint}}}

\newcommand{\clust}{\mbox{\textsc{Cluster}}}

\DeclareUnicodeCharacter{2581}{\underscore}

\title{Improving Multilingual Models with Language-Clustered Vocabularies}

\author{Hyung Won Chung\thanks{ \, Work done as a member of the Google AI Residency Program.} ~~~~~~ Dan Garrette  ~~~~~~ Kiat Chuan Tan ~~~~~~ Jason Riesa \\
  Google Research \\
  \texttt{\{hwchung,dhgarrette,kiatchuan,riesa\}@google.com}
}

\date{}

\begin{document}
\maketitle
\begin{abstract}


State-of-the-art multilingual models depend on vocabularies that cover all of the languages the model will expect to see at inference time, but the standard methods for generating those vocabularies are not ideal for massively multilingual applications. In this work, we introduce a novel procedure for multilingual vocabulary generation that combines the separately trained vocabularies of several automatically derived language clusters, thus balancing the trade-off between cross-lingual subword sharing and language-specific vocabularies. Our experiments show improvements across languages on key multilingual benchmark tasks \tydi{} (+2.9 F1), XNLI (+2.1\%), and WikiAnn NER (+2.8 F1) and factor of 8 reduction in out-of-vocabulary rate, all without increasing the size of the model or data.

\end{abstract}




\section{Introduction}
\label{sec:introduction}

Multilingual models such as mBERT \cite{Devlin2019}, XLM \cite{Lample2019_XLM}, and XLM-R \cite{Conneau2019_XLMR} have built on the advances of deep contextualized language modeling by pretraining on texts from many languages at once.
One trait common to all of these models is the use of a single vocabulary containing subwords from all languages, used to segment the input text before transforming it into a sequence of embeddings. \citet{Conneau2019_XLMR} showed that
increasing the vocabulary size can produce quality gains, but unlike similar monolingual models, the vocabulary embedding matrix in each of these multilingual models constitutes a significant fraction of its total parameters; for example, 47\% of XLM-R's parameters are in its embedding matrix.
Therefore, scaling up a model's vocabulary size
requires the construction of inductive biases that will guide the training procedure to effectively and efficiently learn these parameters.

The multilingual subword vocabularies used by the state-of-the-art models are generated by algorithms such as WordPiece \cite{schuster2012_wpm,Wu2016_WPM}, SentencePiece \cite{Kudo2018_SPM},\footnote{SentencePiece uses BPE or unigram language model.} or Byte Pair Encoding (BPE) \cite{Sennrich2016}. Given a desired vocabulary size, these algorithms select an inventory of subwords that compactly represents the training corpora, which means preferring subwords that occur frequently, and, by extension for multilingual models, occur frequently across languages.



Because these algorithms look at overall subword frequencies in the combined multilingual corpus, they may learn suboptimal decompositions for low-resource languages that happen to have character combinations that resemble subwords of high-resources languages.\footnote{For example, the fact that \textit{la} is a good subword in French and Spanish does not mean that the algorithm should split the prefix \textit{la} away from words in all languages.}
Additionally, subwords in common scripts like Latin and Cyrillic have a higher chance of selection since their counts are combined across a large number of languages \cite{Wu2019}.

By attempting to optimize for the best overall vocabulary across all languages without regard for the differences among those languages, the joint procedure over-emphasizes cross-lingual subword sharing---even across languages with little lexical overlap---which is at odds with the finding of \mbox{\citet{KWMR20}} that subword sharing is not the principal reason for the effectiveness of multilingual models. It also \textit{under}-emphasizes the need for all languages---particularly low-resource languages or those written in scripts used by few languages---to contribute subwords that are most effective for their own representations.

In this paper, we propose a novel approach to multilingual vocabulary generation that seeks to balance the trade-off between optimizing for cross-lingual subword sharing and the need for robust representation of individual languages. At a high level, we: 1)~automatically group languages into clusters based on the distributional similarities of their individual subword vocabularies, 2)~apply the SentencePiece algorithm separately to the data for each individual cluster, and finally 3)~combine all cluster-vocabularies together to form a single unified multilingual model vocabulary.

We evaluate our approach on three distinct downstream tasks:
\tydi{}, XNLI, and WikiAnn NER.
Our experimental results show that our method improves model performance on all three tasks, and achieves a state-of-the-art result for \tydi{} and zero-shot cross-lingual NER. Crucially, our method improves performance without any changes to the model, and since it does not depend on the model architecture, it can be applied to any model that uses a vocabulary.

\section{Clustered Vocabularies}
\label{sec:vocab_generation_process}

\begin{figure}[htb]
  \centering
    \includegraphics[width=6.5cm]{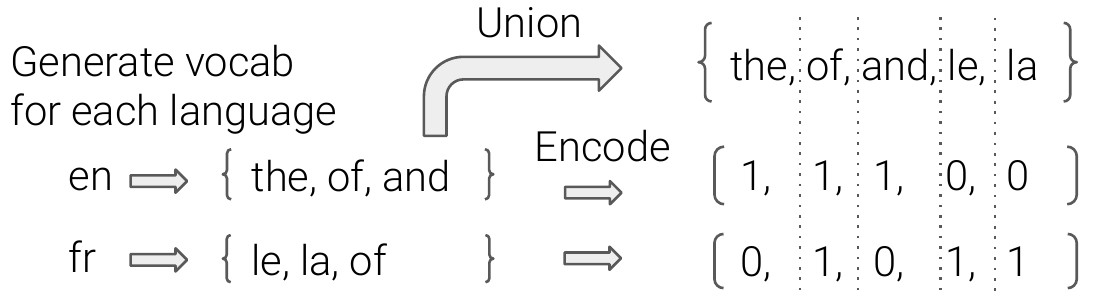}
  \caption{Encoding languages into binary vectors.}
  \label{fig:language_encoding_process}
\end{figure}

We begin by defining a vector representation for each language (Figure~\ref{fig:language_encoding_process}).
For each language $l$ in the set of languages $L$, we generate an $l$-specific vocabulary $V^l$ by running the SentencePiece algorithm on $l$'s corpus. Then we take the union of the resulting vocabulary to form the global vocabulary, $V^{L} = \bigcup_{l \in L} V^l$. Next, for each language, we form a binary vector $\bold{v}^l$ of dimension $\left| V^{L} \right|$, where each component of $\mathbf{v}^l$ corresponds to a subword in $V^{L}$, i.e., $\bold{v}^{l}_{i} = \mathbb{I}[V^{L}_i \in V^{l}]$ where $\mathbb{I}$ is an indicator function. In other words, $\bold{v}^l$ contains a $1$ corresponding to each subword that is present in $l$'s vocabulary, and two languages will have 
similar representations
when their vocabularies have more subwords in common.

With each language encoded as a vector $\bold{v}^l$, we can apply standard clustering algorithms to assign each $l$ to some $c_j$ in the set of clusters $C$. For our experiments, we used $k$-means with cosine distance. Preliminary experiments (Table~\ref{table:values_of_k}) indicated that using $k=8$ would yield good results for our 104-language (the same set of languages used in mBERT) pretraining data so we fixed this value for all remaining experiments.

\begin{table}
\resizebox{7.7cm}{!}{%
\begin{tabular}{ccccccc}
\toprule
$k$ & 1 & 4 & 8 & 16 & 32 & 104 \\
\midrule
\tydi{} & 55.6/69.4 & 56.9/70.5 & \textbf{58.5/71.7} & 57.3/70.9 & 57.0/70.2 & 55.5/69.3 \\
\bottomrule
\end{tabular}}
\caption{\tydi{} results for our $k$-means-based vocabulary generation approach on different values of $k$.  $k=1$ puts all languages in a single cluster, and is thus equivalent to the baseline \joint{} approach; $k=104$ generates a separate vocabulary for each language.}
\label{table:values_of_k}
\end{table}

We then generate a vocabulary $V^{c_j}$ for each cluster $c_j \in C$ by pooling all of the pretraining data for all languages $l \in c_j$ and running the SentencePiece algorithm. We set the target vocabulary size to be proportional to the size of the union of the individual vocabularies $V^l$ of the languages belonging to the cluster i.e. $|V^{c_j}| \propto |\bigcup_{l \in c_j} V^l|$, ensuring that the proportion of the overall vocabulary allocated to a particular cluster is guided by two factors: it will increase if the cluster has more languages, and decrease if the cluster's languages have more vocabulary overlap.
Finally, the multilingual model's overall vocabulary is the union of all of the cluster vocabularies: $V^C = \bigcup_{c_j \in C} V^{c_j}$.

Note that our method is a generalization of the conventional approach, which has $|C| = 1$.


\section{Intrinsic Analysis}
\label{sec:intrinsic_evaluation}

In this section, we directly examine the vocabularies produced by the standard recipe (\joint{}) and our clustering-based method (\clust{}) in order to assess the degree to which each approach is able to capture and balance the inductive biases introduced in \S\ref{sec:introduction}: a multilingual vocabulary should encourage subword sharing across languages when appropriate, but each language should also have the freedom to contribute the subwords that are most effective for its own representation.

For all of our experiments, we generate vocabularies from the Wikipedia articles of 104 languages with 906M sentences. Each multilingual vocabulary is 488k subwords. The cluster assignments and sizes generated by \clust{} are found in Table~\ref{table:cluster_definition}.

\begin{table}[t]
\centering
\resizebox{7.7cm}{!}{%
\begin{tabular}{clr}
\toprule \textbf{Name} & \textbf{Clusters} & \multicolumn{1}{c}{$|V^{c}|$} \\ \midrule
$c_1$ & \makecell[l]{af, sq, hy, az, bn, bs, my, ceb, hr, en, fi, ka, el, he \\ hu, id, ga, jv, lv, lt, ms, min, pl, pa, sco, hbs, sl\\ su, sw, tl, th, tr, uz, vi, cy, yo} & 200,306 \\ \midrule
$c_2$ & ar, bpy, fa, azb, ur, lah & 40,218 \\ \midrule
$c_3$ & \makecell[l]{an, ast, eu, ca, gl, io, it, la, lmo, oc, pms \\ pt, ro, scn, es, war} & 80,764 \\ \midrule
$c_4$ & \makecell[l]{ba, be, bg, ce, cv, kk, ky, mk, mn, ru, sr, tg, tt, uk} & 82,163 \\ \midrule
$c_5$ & bar, br, fr, de, ht, nds, lb, mg, vo & 51,653 \\ \midrule
$c_6$ & zh-Hans, zh-Hant, ja, ko & 25,528 \\ \midrule
$c_7$ & cs, da, nl, et, is, nb, nn, sk, sv, fy & 57,681 \\ \midrule
$c_8$ & gu, hi, kn, ml, mr, ne, new, ta, te & 61,683 \\
\bottomrule
\end{tabular}}
\caption{Cluster definitions of \clust{}.}
\label{table:cluster_definition}
\end{table}


In order to quantify the extent of subword sharing between languages, we look at each language as a \textit{distribution} over the vocabulary. In particular, we apply the multilingual vocabulary $V$'s segmentation to each language $l$'s monolingual corpus in order to count the frequency of each subword. The empirical distribution of a language $l$ over the vocabulary $V$ is defined by normalizing the frequencies of the monolingual corpus. Given these distributional representations of languages, we can use the Wasserstein-1 distance, $W_1$, between two languages to quantify the extent of subword sharing (Table~\ref{table:wasserstein}). Unlike \joint{} for which $W_1$ is relatively small even for languages with distinct scripts (e.g., English and Urdu), \clust{} manifests much larger values for empirically different languages (and hence in different clusters) while having even smaller values for languages in the same cluster (e.g., Japanese and Chinese). This suggests that the clustering based approach not only minimizes the subword sharing when languages are dissimilar but it also puts more emphasis on subword sharing when necessary.


%

\begin{table}
\centering
\resizebox{7.7cm}{!}{%
\begin{tabular}{ccccc}
\toprule
\joint{} & en & ur & ja & zh \\
\midrule
en & 0 & 8 & 9 & 15 \\
ur &  & 0 & 14 & 20 \\
ja &  &  & 0 & 6 \\
\bottomrule
\end{tabular}
\quad
\begin{tabular}{ccccc}
\toprule
\clust{} & en & ur & ja & zh \\
\midrule
en & 0 & 145 & 306 & 307 \\
ur &  & 0 & 165 & 166 \\
ja &  &  & 0 & 1 \\
\bottomrule
\end{tabular}}
\caption{Wasserstein-1 distance ($\times 1000$).}
\label{table:wasserstein}
\end{table}


\begin{table}[t]
\centering
\resizebox{3.5cm}{!}{%
\begin{tabular}{lcc}
\toprule
 & CJK & Arabic \\
\midrule
\joint{}     & 2.8 & 3.4 \\
\clust{}     & 4.4 & 6.1 \\
\bottomrule
\end{tabular}}
\caption{Percentage of each vocabulary's subwords that contain CJK, or Arabic script characters.
}
\label{table:subword_script_fraction}
\end{table}

We quantify the degree of language freedom granted by each approach by examining the fraction of the vocabulary's subwords that contain rare scripts (Table~\ref{table:subword_script_fraction}).
\clust{} has a higher percentage of Chinese, Japanese and Korean (grouped together as CJK) subwords because CJK languages are in the same cluster and by themselves, so CJK subwords can be selected independent of other languages. A similar pattern exists for Arabic script.






\section{Experiments}
\label{sec:experiments}


\begin{table*}
\centering
\resizebox{5in}{!}{%
\begin{tabular}{lcccccccccccc}
\toprule
 & (en) & ar & bn & fi & id & ja & ko & ru & sw & te & th & \avg{} \\
 \midrule
\textbf{\minspan} &&&&&&&&&&& \\
\joint{} & 48.4 & 70.4 & 61.5 & 54.0 & 55.4 & 42.6 & 40.6 & 48.2 & 53.6 & 75.8 & 54.0 & 55.6 \\
\clust{} & \textbf{50.3} & \textbf{71.9} & \textbf{64.8} & \textbf{57.3} & \textbf{58.0} & \textbf{47.2} & \textbf{45.5} & \textbf{49.4} & \textbf{57.0} & \textbf{77.5} & \textbf{56.1} & \textbf{58.5} \\
\textbf{\selectp} &&&&&&&&&&& \\
\joint{} & 63.7 & 83.6 & 71.5 & 65.5 & 65.8 & 55.8 & 63.3 & 67.7 & 66.8 & 84.9 & 69.4 & 69.4 \\
\clust{} & \textbf{65.2} & \textbf{85.7} & \textbf{72.2} & \textbf{68.8} & \textbf{69.6} & \textbf{59.2} & \textbf{65.5} & \textbf{67.9} & \textbf{71.3} & \textbf{86.1} & \textbf{70.7} & \textbf{71.7} \\
\bottomrule
\end{tabular}
}
\caption{Results on the \tydi{} primary tasks: minimal answer span (\minspan) and passage selection (\selectp). The final column~(\avg{}) is the macro average \textit{excluding English}, following \citet{tydiqa}.}
\label{table:tydi}
\end{table*}

\begin{table*}
\centering
\resizebox{\textwidth}{!}{%
\begin{tabular}{lcccccccccccccccc}
\toprule
 & ar & bg & de & el & en & es & fr & hi & ru & sw & th & tr & ur & vi & zh & \avg{} \\
 \midrule
\joint{} & 66.6 & 70.8 & 72.1 & 66.9 & \textbf{81.3} & 74.4 & \textbf{74.1} & 63.6 & 70.8 & 54.1 & 64.6 & 65.4 & 60.6 & 68.3 & 66.4 & 68.0 \\
\clust{} & \textbf{69.1} & \textbf{71.9} & \textbf{72.5} & \textbf{72.6} & \textbf{81.3} & \textbf{75.4} & 74.0 & \textbf{68.0} & \textbf{71.1} & \textbf{56.7} & \textbf{67.6} & \textbf{66.9} & \textbf{63.7} & \textbf{69.6} & \textbf{71.0} & \textbf{70.1} \\
\bottomrule
\end{tabular}}
\caption{XNLI accuracies.}
\label{table:xnli}
\end{table*}

\begin{table*}
\centering
\resizebox{\textwidth}{!}{%
\begin{tabular}{lccccccccccccccccccccc}
\toprule

 &
\multicolumn{4}{c}{$c_1$}  &
\multicolumn{3}{c}{$c_2$}  &
\multicolumn{2}{c}{$c_3$}  &
\multicolumn{2}{c}{$c_4$}  &
\multicolumn{2}{c}{$c_5$}  &
\multicolumn{1}{c}{$c_6$}  &
\multicolumn{2}{c}{$c_7$}  &
\multicolumn{4}{c}{$c_8$}  & 
\multirow{2}{*}{\avg{}}    \\

\cmidrule(lr){2-5}
\cmidrule(lr){6-8}
\cmidrule(lr){9-10}
\cmidrule(lr){11-12}
\cmidrule(lr){13-14}
\cmidrule(lr){15-15}
\cmidrule(lr){16-17}
\cmidrule(lr){18-21}

 & en & sw & yo & he & ar & fa & ur & es & eu & bg & ru & fr & de & ko & nl & et & hi & ml & ta & te & \\
 \midrule
\joint{} & \textbf{84.5} & 64.7 & 54.8 & 52.0 & 48.9 & 36.4 & 32.9 & \textbf{77.4} & \textbf{54.5} & \textbf{80.4} & \textbf{65.5} & 79.9 & 78.3 & 50.7 & \textbf{82.3} & 75.6 & 64.5 & 56.6 & 58.4 & 48.1 & 61.7 \\
\clust{} & 84.1 & \textbf{72.2} & \textbf{62.8} & \textbf{59.2} & \textbf{53.5} & \textbf{51.0} & \textbf{61.8} & 75.5 & 53.1 & 78.6 & 62.0 & \textbf{81.8} & \textbf{78.4} & \textbf{56.8} & 81.8 & \textbf{77.0} & \textbf{70.8} & \textbf{63.2} & \textbf{60.9} & \textbf{56.8} & \textbf{64.5} \\
\bottomrule
\end{tabular}}
\caption{Results for zero-shot NER using cross-lingual transfer from English (following XTREME \cite{Hu2020}) for a sample of languages, grouped according to the clustering used by \clust{}. All scores are labeled span F1, and \avg{} is the macro average across all 40 XTREME languages.}
\label{table:wikiann_group_by_clusters}
\end{table*}


The principal goal of this work is to investigate the effect of improved \emph{vocabulary composition} on multilingual models. We make a good-faith effort to control all other variables, e.g., hyperparameters, training/evaluation procedures. In particular, we keep the number of languages constant, since per-language model capacity is known to affect the performance of multilingual models as shown in~\citet{Conneau2019_XLMR}. In addition, we keep the number of parameters constant\footnote{The only exception is the full scale model in~\S\ref{sec:full-scale_model}.}, including the vocabulary size, since the performance of Transformer-based~\cite{Vaswani2017} models is strongly correlated with number of parameters~\citep{Lepikhin2020,Kaplan2020,Raffel2020,Conneau2019_XLMR,Brown2020}.


In order to demonstrate the effectiveness of our approach across languages and downstream tasks, we evaluate our method on three distinct datasets:
\begin{itemize}
    \item \tydi{} \cite{tydiqa}: question answering in 10 languages. Results in Table~\ref{table:tydi}.
    \item XNLI \cite{Conneau2018_XNLI}: natural language inference in 15 languages. Results in Table~\ref{table:xnli}.
    \item WikiAnn NER \cite{Pan2017}: named entity recognition in 40 languages. Results in Table~\ref{table:wikiann_group_by_clusters}.
\end{itemize}

We pretrain using the masked language modeling (MLM) task on the raw text without applying any language-specific pre-processing.\footnote{See Appendix~\ref{sec:appendix_exp_details} for additional training details.}


At a high level, the experimental results demonstrate that our \clust{} approach to multilingual vocabulary generation improves over the standard \joint{} recipe, increasing the macro average performance across languages for all datasets and tasks. For \tydi{}, we see improvements on all languages. 
For XNLI, \clust{} perform better or equally well in all languages except French. The results on individual languages are more mixed for NER, with \clust{} providing large gains on low-resource or rare-script languages, and small losses on high-resource languages.


In the remainder of this section, we provide more in-depth analyses of these results (\S\ref{sec:analysis}), and show that our approach continues to perform well when used to train a large scale model (\S\ref{sec:full-scale_model}).

\subsection{Analysis}
\label{sec:analysis}

We use the \textit{minimum description length} principle (MDL)~\citep{Rissanen1989} to aid in our analysis. Consider an example where we want to encode data with a codebook. For example, each word in the data can be encoded by a unique sequence of bits, which is referred to as a codeword. The MDL principle favors the codebook with the minimal description length.
Following~\citet{Goldwater_thesis}, we define the description length (DL) as the length of combined codebook and encoded data; that is, the sum of the lengths of all codewords in our codebook plus the length of the encoded data.

We apply this to the setting of learning a subword vocabulary for neural network models. Our codebook is a mapping from a subword to a unique integer index of fixed length, typically 32 bits. Therefore, the description length is equivalent to the sum of the number of unique subwords, i.e. the vocabulary size, and the number of integers of the encoded or tokenized input data. Comparing the description length of two vocabularies is therefore equivalent to comparing the total number of subwords after the input corpus is tokenized by each vocabulary. Without loss of generality, we use the average number of tokens per sentence as an equivalent measure of the description length.



As shown in Table~\ref{table:dl_oov_rate}, \clust{} does indeed reduce the DL of the training corpus, which correlates with the performance improvements we see on downstream tasks.
With smaller DL, the input text is encoded with \textit{longer} subwords, which we might expect to lead to a higher out-of-vocabulary (OOV) rate \cite{Arivazhagan2019}. However, \clust{} has an 8 times smaller OOV rate than \joint{} (Table~\ref{table:dl_oov_rate}).
We believe that this is related to the larger extent of language freedom as evidenced by the particularly large OOV rate reductions observed in rare-script languages. For example, the OOV rate is reduced by a factor of 26 (Korean), 18 (Japanese) and 17 (Chinese).

\begin{table}[]
\centering
\resizebox{5cm}{!}{%
\begin{tabular}{lcc}
\toprule
 & Avg. DL & OOV rate [\%] \\
\midrule
\joint{} & 20.9 & 0.200 \\
\clust{} & \textbf{20.2} &\textbf{0.025} \\
\bottomrule
\end{tabular}}
\caption{Average description length and OOV rate. These are computed on the pretraining data with the same sampling strategies used for pretraining.}
\label{table:dl_oov_rate}
\end{table}

The longer DL of \joint{} means the average subword length is shorter. 
As a result, the model has learn to map from finer-grained input sequences to semantic meaning \cite{Arivazhagan2019}.
As an extreme example, a character-based model would have to learn how to \textit{reconstruct} each word, while a word-based model is exempt from this task.
Though the difference in DL is smaller than this extreme case, the same logic applies and \joint{} must learn a more complex function than \clust{}.

Finally, we note that this correlation between DL and downstream performance means we can use DL as a proxy metric to compare vocabularies, allowing for faster iteration over various vocabulary generation approaches without having to run the expensive model training.


\paragraph{Low-resource languages with common scripts.}
\label{sec:common_script_langs}


Languages like Swahili and Yoruba use Latin script but have small amounts of data and \clust{} outperforms \joint{} for all three tasks on these languages, which we believe is attributable to better segmentation. In \S\ref{sec:introduction}, we highlighted one example where over-segmentation can occur,\footnote{While \joint{} segments the common Swahili adverb \mbox{\textit{lazima}} into \textit{la} and \textit{zima}, \clust{} keeps it intact.} but we can quantify this analysis more generally with DL:
\clust{} has 9.4\% (Swahili) and 11.1\% (Yoruba) shorter DL compared to \joint{}, and this matches well with 7.5 and 8.0 higher NER F1, respectively.


\paragraph{Rare-script languages.}
\label{sec:HR_langs}

For languages with rare scripts (e.g., CJK and Thai), \clust{} strongly outperforms \joint{} in all tasks. For NER, particularly large gains are achieved for Arabic-script languages in cluster $c_2$ (e.g., 28.9 F1 improvement in Urdu) and Indian languages in $c_8$. We hypothesize that the gain for this group of languages is due to the higher coverage of subwords in rare scripts, and consequently lower OOV rates (Table~\ref{table:subword_script_fraction}).

\paragraph{On clusters with languages in different scripts.}
Our vocabularies were trained from the Wikipedia articles, which frequently contain translations/transliterations. For example, the first sentence of both the Hindi- and Tamil-language versions of the article on ``India" contain the exact English phrase ``Republic of India". Since many of the same people/places will be topics in articles across Indic-script languages, it is not too surprising that the clustering algorithm groups some of these languages. We note that the NER performance on the Indic languages in cluster $c_8$ are especially strong (Table~\ref{table:wikiann_group_by_clusters}) possibly due to such shared representation of named entities.

A similar phenomenon can be seen with Korean, since Chinese characters are often appended in parentheses to a Korean word to disambiguate polysemy, which is particularly common for formal contents like Wikipedia. This is why Chinese, Japanese (having large lexical overlap with Chinese) and Korean are in the same cluster $c_6$. We note that these languages show especially strong improvement in all three tasks we considered as well as drastically large reduction in OOV rate.

Therefore, we see these behaviors as a \emph{strength} of our data-driven approach, flexibly capturing the characteristics of the data which may not be obvious from a purely linguistic perspective.

\vspace{5pt}
\subsection{Full-scale model}
\label{sec:full-scale_model}


To evaluate the effectiveness of our method on a large scale model, 
we train a 24-layer Transformer model with \clust{} and report the macro-averaged results in Table~\ref{table:full_scale}. We drastically outperform the baseline for \tydi{} with about 10.7~F1 absolute improvement on \minspan{} and 13.3~F1 on \selectp{}, and NER with 8.2~F1 absolute improvement over XLM-R.  There is a small loss on XNLI, though this may be due to training on less data, and only Wikipedia domain.



\begin{table}
\centering
\resizebox{5.5cm}{!}{%
\begin{tabular}{lccc}
\toprule
\textbf{Model} & \textbf{\tydi{}} & \textbf{XNLI} & \textbf{NER} \\
\midrule
mBERT & 52.7/64.4 & 65.4 & 62.2 \\
XLM-R & - & \textbf{79.2} & 65.4 \\
Ours & \textbf{63.4/77.7} & 77.0 & \textbf{73.6} \\
\midrule
Human & 70.1/79.9 & 92.8 & - \\
\bottomrule
\end{tabular}}
\caption{Comparisons with a full-scale model using our approach. Scores are macro averages over languages. \tydi{} numbers are in (\minspan{}/\selectp{}) format, and our results are on the test set via a submission to the official leaderboard.  Baseline numbers for \tydi{} are from \citet{tydiqa} and the rest are from \citet{Hu2020}.}
\label{table:full_scale}
\end{table}

\vspace{5pt}

\section{Conclusion}

We describe a novel clustering-based multilingual vocabulary generation algorithm. We showed that this empirically motivated clustering-based method consistently outperforms the standard vocabulary generation recipe used by most multilingual pretrained language modeling work without increasing the model size, compute or data.





\section*{Acknowledgements}
We would like to thank
Vera Axelrod,
Tim Dozat,
Melvin Johnson,
Thibault F\'evry,
Xavier Garcia,
and
Karthik Raman
for their feedback on this work.

\bibliographystyle{acl_natbib}
\bibliography{emnlp2020}

\clearpage

\appendix

\section{Experiment details}
\label{sec:appendix_exp_details}

\subsection{Pretraining}

We did not use any hyperparameter search for pretraining. We use LAMB optimizer \cite{You2019_LAMB} with batch size of 4096. \citet{You2019_LAMB} also recommended learning rate of 0.0018, warm-up proportion of 2.5\% of the total number of steps. We use linear warm-up and linear learning rate decay down to 0 in the last step. We use gradient clipping with a norm of 1.0. 

For all the experiments except for the full-scale model, we trained using 64 Google Cloud TPUs. We pretrained models with two different sequence lengths: 128 and 512 with 500k and 125k steps, respectively. The model with sequence length of 512 runs at about 2.3 steps/second, which takes about 16 hours to finish; this model was used to run WikiAnn NER and \tydi{}. The model with sequence length of 128 runs at 9.8 steps/second and finishes in about 15 hours; this model was used to run XNLI. A step refers to one gradient update of the LAMB optimizer.

During pretraining, we sampled each language's data with the following strategy. First we compute the empirical distribution for each language
\begin{equation}
    p^{l} = \frac{n^{l}}{\sum_{l' \in L} n^{l'}}
\end{equation}
where $n^l$ is the number of sentences in language $l$'s corpus. Then we use the exponential smoothing value of 0.7 following \cite{Devlin2019}, i.e., we exponentiate $p^{l}$ and renormalize to compute the sampling probabilities of each language.

We used whole-word masking during pretraining.  However, since some languages do not typically use whitespace between words (e.g., Thai), we used the heuristic of SentencePiece meta symbol U+2581 to designate the beginning of the word. Therefore, a word is defined as the token span between two successive U+2581 symbols. 

For the full-scale model, we pretrained with 256 Google Cloud TPUs for 1.5 million steps with a batch size of 4096 and learning rate of 0.0018, which took 8 days. We only trained one model, and used a sequence length of 512.

All experiments were run in TensorFlow.

\subsection{SentencePiece configurations}
We used the following configuration to train a SentencePiece model: unigram language model, character coverage of 0.9995, and 1M seed sentences.

\subsection{Model architecture}

Except for the full-scale model, all models were 12 layers of Transformers, with hidden size of 768 and 12 attention heads. For faster experimentation, we used an embedding size of 128, similar to \citet{Lan2019}. The total number of parameters is 150M. The full-scale model has 24 Transformer layers, hidden size of 1024, and 16 attention heads. We used an embedding size of 512, totaling 550M parameters. We chose this number of parameters to mimic XLM-R.


\begin{table*}[!htb]
\centering
\resizebox{\textwidth}{!}{%
\begin{tabular}{lcccccccccccccccc}
\toprule
 & ar & bg & de & el & en & es & fr & hi & ru & sw & th & tr & ur & vi & zh & \avg{} \\
 \midrule
\joint & 66.9 & 70.2 & 72.9 & 67.0 & \textbf{81.4} & 75.2 & 74.5 & 63.3 & \textbf{71.0} & 53.5 & 63.0 & 66.1 & 60.9 & 68.5 & 67.4 & 68.1 \\
\clust{} & \textbf{68.3} & \textbf{71.7} & \textbf{73.5} & \textbf{74.1} & 81.0 & \textbf{75.4} & \textbf{75.3} & \textbf{67.4} & 70.6 & \textbf{57.1} & \textbf{67.8} & \textbf{66.9} & \textbf{62.9} & \textbf{69.4} & \textbf{70.9} & \textbf{70.2} \\

\bottomrule
\end{tabular}}
\caption{XNLI accuracy on development set. The best performing hyperparameters for both \joint{} and \clust{} were learning rate of $2 \times 10^{-5}$, batch size of 32, and 3 training epochs.}
\label{table:xnli_dev}
\end{table*}

\begin{table*}
\centering
\resizebox{\textwidth}{!}{%
\begin{tabular}{lccccccccccccccccccccc}
\toprule

 &
\multicolumn{4}{c}{$c_1$}  &
\multicolumn{3}{c}{$c_2$}  &
\multicolumn{2}{c}{$c_3$}  &
\multicolumn{2}{c}{$c_4$}  &
\multicolumn{2}{c}{$c_5$}  &
\multicolumn{1}{c}{$c_6$}  &
\multicolumn{2}{c}{$c_7$}  &
\multicolumn{4}{c}{$c_8$}  & 
\multirow{2}{*}{\avg{}}    \\

\cmidrule(lr){2-5}
\cmidrule(lr){6-8}
\cmidrule(lr){9-10}
\cmidrule(lr){11-12}
\cmidrule(lr){13-14}
\cmidrule(lr){15-15}
\cmidrule(lr){16-17}
\cmidrule(lr){18-21}

 & en & sw & yo & he & ar & fa & ur & es & eu & bg & ru & fr & de & ko & nl & et & hi & ml & ta & te &  \\
 \hline
\joint{} & \textbf{84.3} & 64.8 & 51.9 & 51.7 & 49.0 & 36.3 & 31.7 & \textbf{76.7} & \textbf{53.3} & \textbf{81.0} & \textbf{66.4} & 79.9 & 78.2 & 51.1 & \textbf{81.9} & 76.0 & 63.4 & 53.4 & 59.2 & 49.1 & 61.3 \\
\clust{} & 84.1 & \textbf{72.1} & \textbf{64.1} & \textbf{58.8} & \textbf{54.2} & \textbf{50.9} & \textbf{61.8} & 74.7 & 51.8 & 79.3 & 63.0 & \textbf{81.5} & \textbf{78.6} & \textbf{57.3} & 81.2 & \textbf{76.9} & \textbf{72.1} & \textbf{61.4} & \textbf{61.9} & \textbf{56.0} & \textbf{64.3} \\
\hline
\end{tabular}}
\caption{WikiAnn NER F1 scores on development set. We used $4\times 10^{-5}$, batch size of 32 and 2 training epochs for both \clust{} and \joint{}.}
\label{table:wikiann_group_by_clusters_dev}
\end{table*}


\begin{table*}[t]
\centering
\resizebox{13cm}{!}{%
\begin{tabular}{lccccccccccc}
\toprule
 & ar & bn & fi & id & ja & ko & ru & sw & te & th & \avg{} \\
\midrule
\textbf{\minspan} & 77.0 & 72.1 & 66.1 & 65.1 & 52.9 & 55.8 & 57.2 & 65.9 & 82.3 & 59.7 & 65.4 \\
\textbf{\selectp} & 88.5 & 81.3 & 76.7 & 76.0 & 66.1 & 72.7 & 74.3 & 80.1 & 90.6 & 73.8 & 78.0 \\
\bottomrule
\end{tabular}
}
\caption{Full-scale model's result on the \tydi{} primary tasks (development set). }
\label{table:tydi_full_scale_dev}
\end{table*}

\begin{table*}[t]
\centering
\resizebox{\textwidth}{!}{%
\begin{tabular}{cccccccccccccccc}
\toprule
ar & bg & de & el & en & es & fr & hi & ru & sw & th & tr & ur & vi & zh & \avg{} \\
\midrule
77.1 & 79.9 & 81.4 & 80.9 & 86.9 & 82.9 & 82.0 & 75.5 & 79.2 & 59.4 & 74.4 & 73.5 & 69.2 & 77.8 & 77.2 & 77.2 \\
\bottomrule
\end{tabular}}
\caption{XNLI (development set) result for the full-scale model. The best performing model has learning rate of 0.0001, train epochs of 9 and batch size of 512.}
\label{table:xnli_dev_full_scale}
\end{table*}

\begin{table*}[h]
\centering
\resizebox{\textwidth}{!}{%
\begin{tabular}{lccccccccccccccccccccc}
\toprule

 &
\multicolumn{4}{c}{$c_1$}  &
\multicolumn{3}{c}{$c_2$}  &
\multicolumn{2}{c}{$c_3$}  &
\multicolumn{2}{c}{$c_4$}  &
\multicolumn{2}{c}{$c_5$}  &
\multicolumn{1}{c}{$c_6$}  &
\multicolumn{2}{c}{$c_7$}  &
\multicolumn{4}{c}{$c_8$}  & 
\multirow{2}{*}{\avg{}}    \\

\cmidrule(lr){2-5}
\cmidrule(lr){6-8}
\cmidrule(lr){9-10}
\cmidrule(lr){11-12}
\cmidrule(lr){13-14}
\cmidrule(lr){15-15}
\cmidrule(lr){16-17}
\cmidrule(lr){18-21}

 & en & sw & yo & he & ar & fa & ur & es & eu & bg & ru & fr & de & ko & nl & et & hi & ml & ta & te & \\
 \hline
\clust{} & 86.4 & 73.4 & 79.1 & 72.6 & 73.2 & 75.4 & 82.3 & 85.7 & 68.2 & 88.4 & 79.8 & 87.8 & 84.0 & 68.2 & 87.4 & 85.3 & 79.4 & 73.9 & 73.2 & 67.2 & 73.7 \\
\hline
\end{tabular}}
\caption{WikiAnn NER F1 scores of the full-scale model on development set. The best performing model has learning rate of $3 \times 10^{-5}$, batch size of 32 and trained for 2 epochs.}
\label{table:wikiann_full_scale_group_by_clusters_dev}
\end{table*}


\subsection{Fine-tuning}

We ran experiments with two seed values and chose the best model based on the average of the two runs. For fine-tuning, we used Adam optimizer \cite{Kingma2015}.

For WikiAnn NER, we used a learning rate of $4 \times 10^{-5}$, a batch size of 32, and 2 training epochs. We found that the performance is robust with respect to the set of hyperparameters, so we did not change this setting. The training was run with 4 TPUs which took about one hour to finish. The evaluation metric is span-level F1 score. Our evaluation code was tested against the \texttt{seqeval} library: \href{https://github.com/chakki-works/seqeval}{https://github.com/chakki-works/seqeval}, and produces the same scores.

For XNLI, we used a batch size of 32, performed grid search over the learning rate of $[1\times 10^{-5},\ 2\times 10^{-5}, \ 3 \times 10^{-5}]$, and trained for 3 epochs. We chose the best model on the development set based on the macro-averaged accuracy and then used that model to report on the test set. The training was run with 8 TPUs which took about 2-3 hours to finish. The evaluation metric is classification accuracy.

For \tydi{}, we found that larger batch sizes improved the training stability, so we used a batch size of 512. With the larger batch size, longer epochs were helpful, so we used a grid search over the learning rate of $[3\times 10^{-5}, \ 4\times 10^{-5}, \ 5 \times 10^{-5}]$ and training epochs of $[7, \  8, \  9]$. We chose the best model based on the macro-averaged F1 score on 10 languages, excluding English following \citet{tydiqa}. For the hyperparameters that are specific to \tydi{}, we used the same settings as the baseline model from \cite{tydiqa}: 45 maximum passages, 0.1 include unknown rates, sequence length of 512, and window stride of 128. The evaluation metric is F1 score, which we computed with the official evaluation script from \href{https://github.com/google-research-datasets/tydiqa}{https://github.com/google-research-datasets/tydiqa}.

\begin{table*}[t]
\centering
\resizebox{7.7cm}{!}{%
\begin{tabular}{lccc}
\toprule
\multicolumn{1}{l}{} & Training & Development & Test \\
\midrule
XNLI & 392,702 & 2490 & 5010 \\
\tydi{} & 166916 & 18670 & 18751 \\
WikiAnn NER & 20000 & 262300 & 262300 \\
\bottomrule
\end{tabular}}
\caption{Number of examples in training, development, test set splits for each evaluation datasets.}
\label{table:num_examples}
\end{table*}

\section{Datasets}

For pretraining, we use 906M sentences of Wikipedia data covering 104 languages. The number of examples of the three datasets used for evaluation is summarized in Table~\ref{table:num_examples}. The XNLI data has a training set in English and development and test sets in 15 languages, which can be downloaded from \href{https://cims.nyu.edu/~sbowman/multinli/}{https://cims.nyu.edu/~sbowman/multinli/}.

The \tydi{} datasets are in 11 languages including English, which is excluded from the official evaluation. The link to download the dataset is \href{https://github.com/google-research-datasets/tydiqa#download-the-dataset}{https://github.com/google-research-datasets/tydiqa}.

For Wikiann NER data, we follow XTREME \cite{Hu2020} and used the balanced train, development, test set splits of \citet{Rahimi2019}, for 40 languages. The dataset can be downloaded from \href{https://github.com/afshinrahimi/mmner}{https://github.com/afshinrahimi/mmner}.  

We did not exclude any examples from the three datasets. For Wikiann NER, Greek, Thai, Japanese, Korean and Chinese data have incorrect IOB2 encoding, e.g., \texttt{I-PER} following \texttt{O}. We fixed those encoding with a simple rule such that a tag with \texttt{I} prefix starting after \texttt{O} is corrected to have \texttt{B} prefix.

We did not use any preprocessing for XNLI or \tydi{}.

\section{Performance on development set}

The main body of the paper contains test set results for XNLI and WikiAnn NER. In this section, we report the development set results so that researchers can try to reproduce the results without consulting to the test set.

Table~\ref{table:xnli_dev} shows the results for XNLI and the results for WikiAnn NER are summarized in Table~\ref{table:wikiann_group_by_clusters_dev}. Both tables contain the best performing hyperparameters in the caption. Since the \tydi{} test set is private, we performed all experiments on the development set for \tydi{} except for the full-scale model for which we submitted to the official leaderboard.

\subsection{Development set results for the full-scale model}

For the full-scale model, Table~\ref{table:wikiann_full_scale_group_by_clusters_dev} shows the development set results on WikiAnn NER, Table~\ref{table:xnli_dev_full_scale} for XNLI and, Table~\ref{table:tydi_full_scale_dev} for \tydi{}. We found that this large Transformer model requires different sets of hyperparameters to be effective. We used the LAMB optimizer to match the pretraining since it made the training more stable. For XNLI we did a grid search on over learning rates $[9\times 10^{-5}, \ 1\times 10^{-4}]$, training epochs $[9,\ 10,\ 11]$, and a fixed batch size of 512. For \tydi{} we did a grid search over learning rates $[8\times 10^{-5}, \ 9\times 10^{-5}]$ and training epochs $[13, \ 14,\ 15]$, and a fixed batch size of 512. 
For WikiAnn NER, we chose the learning rate from $[2\times 10^{-5}, \ 3 \times 10^{-5}]$, and used a fixed batch size of 32 and 2 training epochs.

\subsection{WikiAnn NER test set results on all 40 languages}
In this section, we expand the results in and Table~\ref{table:wikiann_group_by_clusters} and Table~\ref{table:full_scale}, which only show subset of languages (or only average for the latter.

\begin{table*}[]
\resizebox{\textwidth}{!}{%
\begin{tabular}{cccccccccccccccccccc}
\toprule
en & af & ar & bg & bn & de & el & es & et & eu & fa & fi & fr & he & hi & hu & id & it & ja & jv \\
\midrule
86.2 & 84.0 & 73.1 & 88.1 & 84.0 & 84.0 & 82.8 & 86.7 & 85.3 & 68.8 & 76.2 & 84.5 & 87.5 & 72.9 & 77.7 & 83.0 & 65.1 & 85.0 & 24.9 & 68.6 \\
\bottomrule
\end{tabular}
}
\vspace{10pt}

\resizebox{\textwidth}{!}{%
\begin{tabular}{cccccccccccccccccccc}
\toprule
ka & kk & ko & ml & mr & ms & my & nl & pt & ru & sw & ta & te & th & tl & tr & ur & vi & yo & zh \\
\midrule
78.8 & 54.0 & 68.0 & 73.5 & 73.8 & 78.0 & 66.4 & 87.6 & 86.5 & 79.5 & 72.7 & 72.1 & 66.0 & 2.3 & 80.8 & 83.3 & 78.6 & 79.1 & 80.2 & 34.1 \\
\bottomrule
\end{tabular}
}
\caption{WikiAnn NER results in Table~\ref{table:wikiann_group_by_clusters} on all 40 languages. The average F1 scores are 61.7 and 64.5 for \joint{} and \clust{}, respectively.}
\label{table:wikiann_all_lang}
\end{table*}
For these test set results, Table~\ref{table:wikiann_all_lang} expands on Table~\ref{table:wikiann_group_by_clusters} and show results on all 40 languages considered in XTREME. Table~\ref{table:full_scale_wikiann_all_lang} expands on Table~\ref{table:full_scale} in a similar manner.

\begin{table*}[]
\resizebox{\textwidth}{!}{%
\begin{tabular}{cccccccccccccccccccc}
\toprule
en & af & ar & bg & bn & de & el & es & et & eu & fa & fi & fr & he & hi & hu & id & it & ja & jv \\
\midrule
86.2 & 84.0 & 73.1 & 88.1 & 84.0 & 84.0 & 82.8 & 86.7 & 85.3 & 68.8 & 76.2 & 84.5 & 87.5 & 72.9 & 77.7 & 83.0 & 65.1 & 85.0 & 24.9 & 68.6 \\
\bottomrule
\end{tabular}
}
\vspace{10pt}

\resizebox{\textwidth}{!}{%
\begin{tabular}{cccccccccccccccccccc}
\toprule
ka & kk & ko & ml & mr & ms & my & nl & pt & ru & sw & ta & te & th & tl & tr & ur & vi & yo & zh \\
\midrule
78.8 & 54.0 & 68.0 & 73.5 & 73.8 & 78.0 & 66.4 & 87.6 & 86.5 & 79.5 & 72.7 & 72.1 & 66.0 & 2.3 & 80.8 & 83.3 & 78.6 & 79.1 & 80.2 & 34.1 \\
\bottomrule
\end{tabular}
}
\caption{WikiAnn NER results in Table~\ref{table:full_scale} on all 40 languages. The average F1 score is 73.6.}
\label{table:full_scale_wikiann_all_lang}
\end{table*}

\subsection{Language information}

Table~\ref{table:list_of_languages} lists all 104 languages considered in this paper and their script information.

\begin{table*}[]
\centering
\resizebox{\textwidth}{!}{%
\begin{tabular}{llr}
\toprule
Language & Language code & Script \\
\midrule
Afrikaans & af & Latin \\
Albanian & sq & Latin \\
Arabic & ar & Arabic \\
Aragonese & an & Latin \\
Armenian & hy & Armenian \\
Asturian & ast & Latin \\
Azerbaijani & az & Latin \\
Bashkir & ba & Cyrillic \\
Basque & eu & Latin \\
Bavarian & bar & Latin \\
Belarusian & be & Cyrillic \\
Bengali & bn & Bengali \\
Bishnupriya-manipuri & bpy & Bengali \\
Bosnian & bs & Latin \\
Breton & br & Latin \\
Bulgarian & bg & Cyrillic \\
Burmese & my & Myanmar \\
Catalan & ca & Latin \\
Cebuano & ceb & Latin \\
Chechen & ce & Cyrillic \\
Chinese-simplified & zh-Hans & Chinese \\
Chinese-traditional & zh-Hant & Chinese \\
Chuvash & cv & Cyrillic \\
Croatian & hr & Latin \\
Czech & cs & Latin \\
Danish & da & Latin \\
Dutch & nl & Latin \\
English & en & Latin \\
Estonian & et & Latin \\
Finnish & fi & Latin \\
French & fr & Latin \\
Galician & gl & Latin \\
Georgian & ka & Georgian \\
German & de & Latin \\
Greek & el & Greek \\
Gujarati & gu & Gujarati \\
Haitian & ht & Latin \\
Hebrew & he & Hebrew \\
Hindi & hi & Devanagari \\
Hungarian & hu & Latin \\
Icelandic & is & Latin \\
Ido & io & Latin \\
Indonesian & id & Latin \\
Irish & ga & Latin \\
Italian & it & Latin \\
Japanese & ja & Japanese \\
Javanese & jv & Latin \\
Kannada & kn & Kannada \\
Kazakh & kk & Cyrillic \\
Kirghiz & ky & Cyrillic \\
Korean & ko & Korean \\
Latin & la & Latin \\
\bottomrule
\end{tabular}

\quad

\begin{tabular}{llr}
\toprule
Language & Language code & Script \\
\midrule
Latvian & lv & Latin \\
Lithuanian & lt & Latin \\
Lombard & lmo & Latin \\
Low-saxon & nds & Latin \\
Luxembourgish & lb & Latin \\
Macedonian & mk & Cyrillic \\
Malagasy & mg & Latin \\
Malay & ms & Latin \\
Malayalam & ml & Malayalam \\
Marathi & mr & Devanagari \\
Minangkabau & min & Latin \\
Mongolian & mn & Cyrillic \\
Nepali & ne & Devanagari \\
Newar & new & Devanagari \\
Norwegian-bokmal & nb & Latin \\
Norwegian-nynorsk & nn & Latin \\
Occitan & oc & Latin \\
Persian & fa & Arabic \\
Piedmontese & pms & Latin \\
Polish & pl & Latin \\
Portuguese & pt & Latin \\
Punjabi & pa & Gurmukhi \\
Romanian & ro & Latin \\
Russian & ru & Cyrillic \\
Scots & sco & Latin \\
Serbian & sr & Cyrillic \\
Serbo-croatian & hbs & Latin \\
Sicilian & scn & Latin \\
Slovak & sk & Latin \\
Slovenian & sl & Latin \\
South-azerbaijani & azb & Arabic \\
Spanish & es & Latin \\
Sundanese & su & Latin \\
Swahili & sw & Latin \\
Swedish & sv & Latin \\
Tagalog & tl & Latin \\
Tajik & tg & Cyrillic \\
Tamil & ta & Tamil \\
Tatar & tt & Cyrillic \\
Telugu & te & Telugu \\
Thai & th & Thai \\
Turkish & tr & Latin \\
Ukrainian & uk & Cyrillic \\
Urdu & ur & Arabic \\
Uzbek & uz & Latin \\
Vietnamese & vi & Latin \\
Volapuk & vo & Latin \\
Waray-waray & war & Latin \\
Welsh & cy & Latin \\
West & fy & Latin \\
Western-punjabi & lah & Arabic \\
Yoruba & yo & Latin \\
\bottomrule
\end{tabular}
}
\caption{List of languages used in the pre-training.}
\label{table:list_of_languages}
\end{table*}

\end{document}